\documentclass[9pt]{article} 
\usepackage{spconf,amsmath,graphicx}
\usepackage{times}
\usepackage{hyperref}
\usepackage{url}
\usepackage[utf8]{inputenc}
\usepackage{graphicx}
\usepackage{tikz}
\usepackage[font=small]{caption}
\usepackage{subcaption}
\usepackage{booktabs}
\usepackage{lineno}
\usepackage{hyperref}

\usetikzlibrary{arrows}
\usetikzlibrary{calc}
\newcommand{\bs}[1]{\boldsymbol{#1}}

\everydisplay{
  \abovedisplayskip=.20\baselineskip  plus.2ex minus.2ex
  \abovedisplayshortskip=0.15\baselineskip plus.2ex minus.2ex
  \belowdisplayskip=0.20\baselineskip plus.2ex minus.2ex
  \belowdisplayshortskip=0.15\baselineskip plus.2ex minus.2ex
}

\title{Towards end-to-end spoken language understanding}

\name{
  Dmitriy Serdyuk
  \sthanks{This work was done when the first author was an intern with Facebook.}$^{1,2}$
  \thinspace Yongqiang Wang$^{1}$ 
  \thinspace Christian Fuegen$^{1}$ 
  \thinspace Anuj Kumar$^{1}$ 
  \thinspace Baiyang Liu$^{1}$
  \thinspace Yoshua Bengio$^{2}$
}

\address{\begin{tabular}{cc}$^{1}$ Facebook & $^{2}$ Universit\'e de Montr\'eal, MILA \\
	One Hacker Way  & 2920 Chemin de la Tour, office 3353,\\
	Menlo Park, CA 94025, USA &  Montreal, QC H3T 1J4, Canada \\
        \end{tabular}\\
        {\em \{serdyuk, yqw, fuegen,anujk,baiyangliu\}@fb.com}
      }

\begin{document}

\tikzstyle{block}=[draw, fill=blue!20, text width=5em, 
    text centered, minimum height=2.5em]
\tikzstyle{block2}=[draw, fill=orange!20, text width=5em, 
    text centered, minimum height=2.5em]

\maketitle
\begin{abstract}
  Spoken language understanding system is traditionally designed as a pipeline of
  a number of components. First, the audio signal is processed
  by an automatic speech recognizer for transcription or n-best hypotheses. With
  the recognition results, a natural language understanding system classifies
  the text to structured data as domain, intent and slots for down-streaming
  consumers, such as dialog system, hands-free applications. These components
  are usually developed and optimized independently. In this paper, we present
  our study on an end-to-end learning system for spoken language understanding. 
  With this unified
  approach, we can infer the semantic meaning directly from audio features
  without the intermediate text representation. 
  This study showed that the trained model can achieve reasonable good result
  and demonstrated that the model can capture the semantic attention directly
  from the audio features.
\end{abstract}

\begin{keywords}
Spoken language understanding, end-to-end training, recurrent neural networks
\end{keywords}

\section{Introduction}

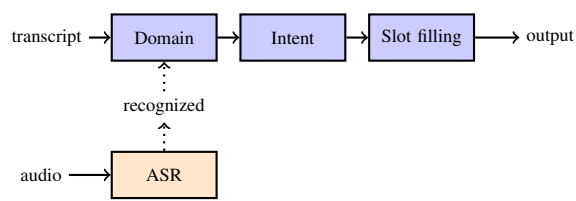
\begin{figure}[tbp]
    \centering
  \begin{tikzpicture}[->,thick]
\scriptsize
\tikzstyle{main}=[circle, minimum size = 7mm, thick, draw =black!80, node distance = 12mm]

\node (transcript) {transcript};
\path (transcript.east)+(1cm,0) node (domain) [block] {Domain}
      (domain.east)+(1cm,0) node (intent) [block] {Intent}
      (intent.east)+(1cm,0) node (slot) [block] {Slot filling}
      (slot.east)+(1cm,0) node (out) {output};
      
\path [draw,->] (transcript) -- (domain);
\path [draw,->] (domain) -- (intent);
\path [draw,->] (intent) -- (slot);
\path [draw,->] (slot) -- (out);

\path (domain.south)+(0, -0.6cm) node (recognized)  {recognized};
\path (recognized.south)+(0, -0.7cm) node (asr) [block2] {ASR};
\path (recognized.south west)+(-1cm, -0.7cm) node (audio)  {audio};

\path [draw,->] (audio) -- (asr);

\path [draw,->,dotted] (asr) -- (recognized);
\path [draw,->,dotted] (recognized.north) -- (domain.south);

\end{tikzpicture}    
    \caption{Traditional SLU system. The NLU part (in blue) is trained on transcripts and consists of a domain classifier,
        intent classifier and a slot filling model. The speech recognition part
        (in orange) is trained independently. During the evaluation time the NLU system uses the
        recognized text from the ASR system, as denoted by the dotted arrows.}
    \label{fig:traditional-slu}
    \vspace{-1.5em}
\end{figure}

With the growing demand of voice interfaces for mobile and virtual reality (VR)
devices, spoken language understanding (SLU) has received many researchers'
attention recently~\cite{tur2011spoken, xu2014contextual, yao2013recurrent,
bhargava2013easy, ravuri2015recurrent, sarikaya2014application, tur2012towards}.
Given a spoken utterance, a typical SLU system performs three main tasks: 
domain classification, intent detection and slot filling~\cite{tur2011spoken}. 
Standard SLU systems are usually
designed as a pipeline structure. As shown in Fig.~\ref{fig:traditional-slu}, 
recorded speech signals are converted
by an automatic speech recognition (ASR) module into the spoken format text, 
followed 
by an optional inverse text normalization module to translate the spoken
domain text to the written domain; then a natural language understanding (NLU)
module is used to determine intent and extract slots accordingly. Although there
are some works (e.g.~\cite{morbini2012reranking}) that take the possible ASR errors into consideration when
designing the NLU module, the pipeline approach is widely adopted. 
One arguable
limitation of this pipelined architecture is that each module is optimized separately
under different criteria: the ASR module is trained to minimize the word error rate
(WER) criterion, which typically weights each word equally; obviously not every word
has the same impact on the intent classification or slot filling accuracy. On the
other hand, the NLU module is typically trained on the clean text (transcription)
without ASR
errors, but during evaluation, it has to use recognized hypotheses outputted by the ASR module
as the input: errors in ASR module, especially in noisy conditions, 
will be propagated to SLU degrading its
performance. 
Second consideration is how human process speech signal to extract intent
level concepts.
Intuitively, when asked to perform intent
detection tasks, humans do not understand speech by
recognizing word by word; instead, humans interpret and understand speech
directly, while attention is given to the high level concepts which are
directly related with the task. 
It is thus desirable to train an SLU system in an
end-to-end fashion.

End-to-end learning has been widely used in several areas, such as machine
translation~\cite{sutskever2014sequence, bahdanau2014neural,
gehring2017convolutional} and
image-captioning\cite{xu2015show}. It has also been 
investigated for speech synthesis \cite{oord2016wavenet} and 
ASR tasks~\cite{amodei2016deep, chan2016listen, soltau2016neural}. 
For example, the CTC loss function is used to
train an ASR system to map the feature sequences directly to the word sequences
in \cite{soltau2016neural} and
it has been shown to perform similarly to the traditional ASR systems; 
in \cite{chorowski2015attention, bahdanau2016end}, 
encode-decoder models with attention have been used for ASR tasks, including
large vocabulary ASR. End-of-end
learning of memory networks is also used for knowledge carryover in multi-turn
spoken language understanding~\cite{chen2016end}.

Inspired by these success, we
explore the possibility to extend the end-to-end ASR learning to include NLU
component and optimize the whole system for SLU purpose. As the first step
towards an end-to-end SLU system, in this work, we focus on maximizing the
single-turn intent
classification accuracy using log-Mel filterbank feature directly. Contributions of
this work are following: 1) we demonstrate the possibility of training a
language understanding model from audio features (see
Section~\ref{sec:end-to-end}). To the best knowledge of authors this is the
first work on this topic; 2) We show that the performance of an SLU degrades when evaluating on the ASR output. The details of our experiments are
    described in Section~\ref{sec:experiments}, and results in Section~\ref{sec:results}; Additionally, we perform experiments on noise-corrupted data. We artificially 
    add noise to the audio data to demonstrate the degradation of the performance of a standard SLU evaluation and test our system.

\section{Standard ASR and NLU systems}
\label{sec:baseline}

\begin{figure}[ht]
    \centering
    \includegraphics[width=0.7\linewidth]{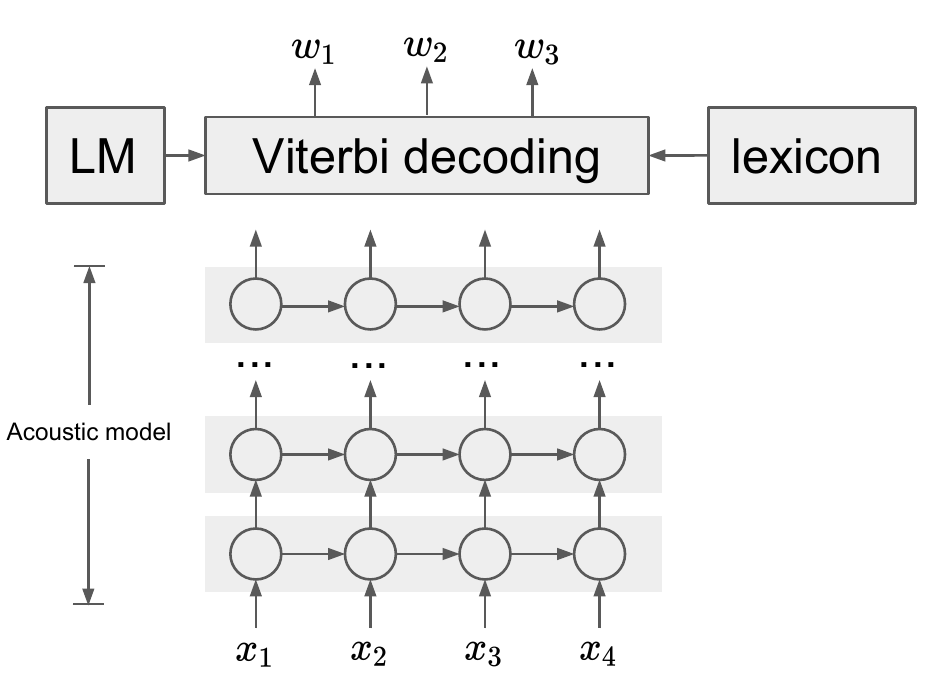}
    \caption{A schematic picture of a traditional ASR system. The recurrent acoustic model
      predicts the states of an HMM. It is decoded with Viterbi algorithm using the language
      model (LM).}
    \label{fig:asr}
\end{figure}

Given a sequence of feature vectors $\bs X = (\bs x_1, \dots, \bs x_T)$, an ASR system is
trained
to find the most likely word sequences $\bs W^{*} = (w_1, \dots, w_n)$ using the chain rule: 
\begin{align*}
  \bs W^{*} = \mbox{arg}\max_{\bs W} p (\bs W | \bs X) = \mbox{arg}\max_{\bs W}
  p (\bs X | \bs W) p(\bs W).
\end{align*}
Therefore, the ASR system is usually divided into two models: an acoustic model
$p(\bs X | \bs W)$ and a language model $p(\bs W)$ (AM and LM respectively).
CD-HMM-LSTM~\cite{sak2014long} is widely used as an AM, in which the feature vector
sequence is converted to the likelihood vectors of context-dependent HMM
states for each
acoustic frame. Together with the LM ($p(\bs W)$ is usually a statistical $n$-gram
model) and a dictionary, a Viterbi decoder is used to search for the most likely
word sequence. Fig.~\ref{fig:asr} depicts the described standard ASR
architecture: the core part of the AM is a multi-layer LSTM~\cite{hochreiter1997long} network, which
predicts the probability of CD-HMM states for each frame. Since most of the SLU 
systems requires spontaneous response, usually only uni-directional
LSTM is used.

\begin{figure}[tbp]
    \centering
    \includegraphics[width=0.8\linewidth]{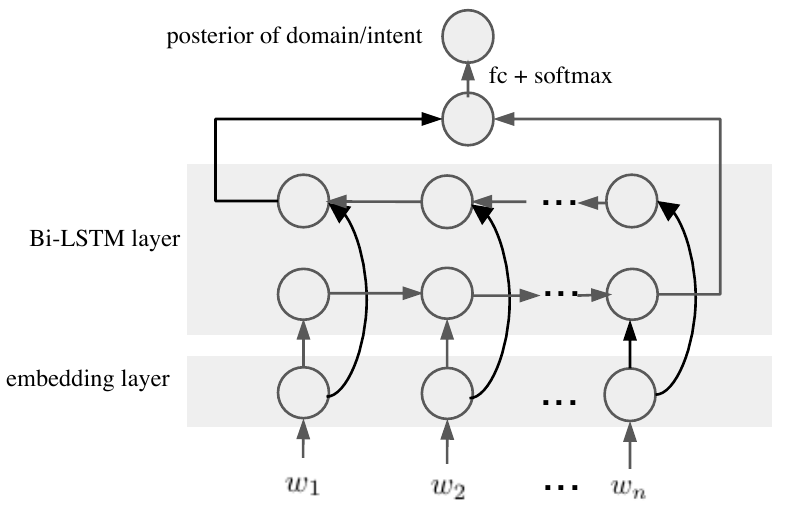}
    \caption{A schematic diagram of a standard NLU system for domain/intent
      classification. ``{\tt fc}'' means a fully-connected layer. }
    \label{fig:nlu}
\end{figure}

Given the word sequence output by the ASR module, an NLU module is used to
perform domain and intent classification, and to fill slots for different
intents accordingly. Following~\cite{ravuri2015recurrent}, we use LSTM-based
utterance classifier, in which the input words are first embedded in a
dense representation, and then the LSTM network is used to encode the word
sequence.  We found 2-layer 
bi-directional LSTM encoder performs better. Fig.~\ref{fig:nlu} demonstrates
how a standard NLU system predicts the domain/intent for a given word sequence. 
Note that since the NLU system incurs much smaller latency compared with the ASR
system and the classification only starts after the entire word sequence become
available, it is possible to use bi-directional RNNs for NLU.

In the pipelined approach to SLU, ASR, and NLU modules are usually trained
independently, where the NLU module is trained using human transcription as the
input. During the evaluation phase, the ASR output is piped into the NLU module.

\section{End-to-end Spoken Language Understanding}
\label{sec:end-to-end}
\begin{figure}[tbp]
    \centering
    \includegraphics[width=1.0\linewidth]{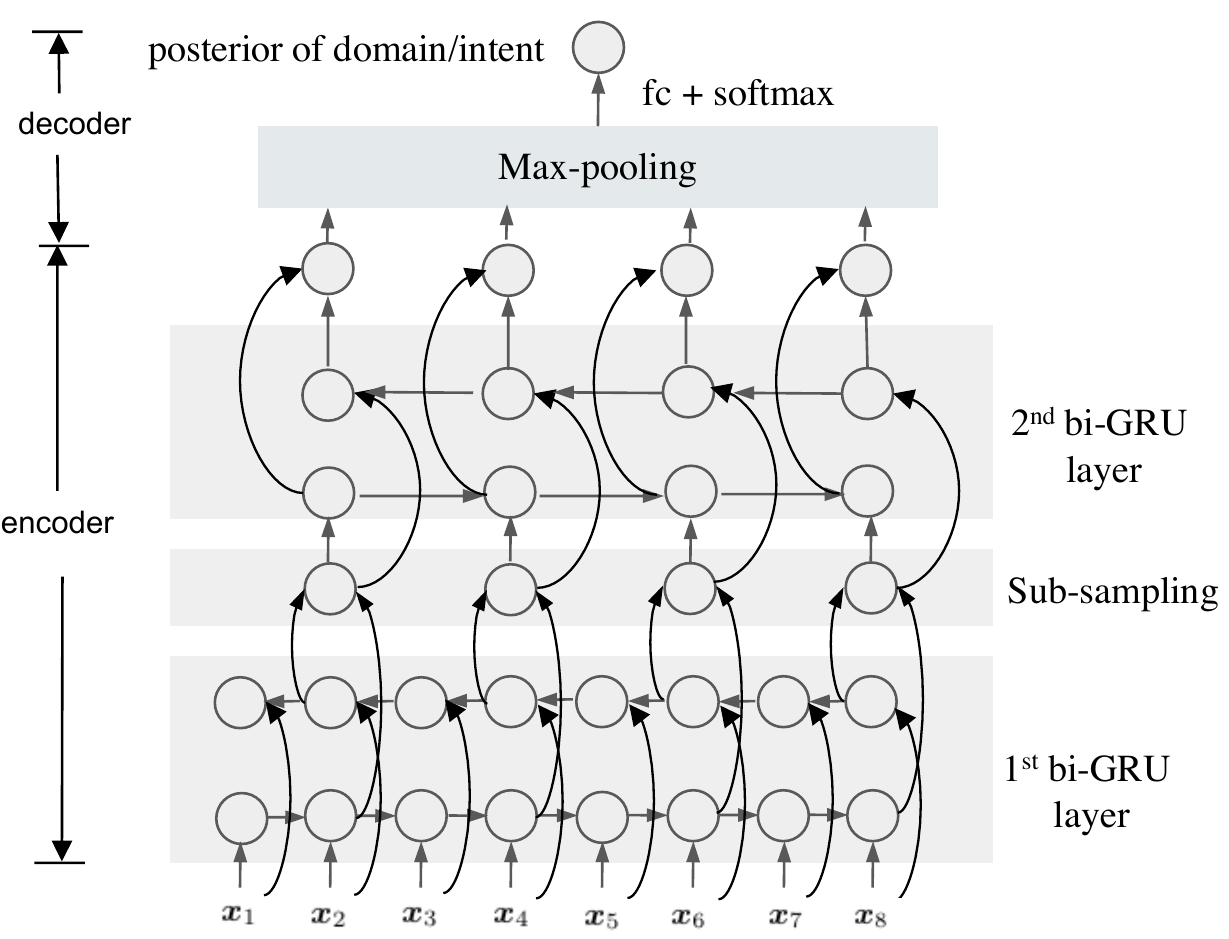}
    \caption{End-to-end SLU system. Sub-sampling is performed at every layer to
    reduce the output sequence length. Our best model uses 4 layer bidirectional GRU.}
    \label{fig:slu}
\end{figure}

As the first step towards the end-to-end spoken language understanding, we focus on two tasks: speech-to-domain and
speech-to-intent. Both tasks are sequence classification problems where the
input is log-Mel filterbank feature vectors.

The task of end-to-end SLU is close to speech recognition with a difference that the structure of the output is simpler but the transformation
is more complicated and the data is noisy and non-unimodal (close utterances may have sufficiently different intents). For this reason our model 
was inspired by works in end-to-end speech recognition~\cite{graves2012sequence, bahdanau2014neural, chan2016listen, battenberg2017exploring}. 
We use an encoder-decoder framework. 
The input log-Mel filterbank features are processed
by an encoder which is a multi-layer bidirectional 
\emph{gated recurrent unit} (GRU,~\cite{chung2014empirical})
network. A potential issue of using log-Mel filterbank feature is that it is
generated every 10 ms: the 10-ms frame rate is good for classifying sub-phone unit
like CD-HMM states, while it may not be suitable for utterance-level
classification as GRUs may forget most of the speech content when it arrives at 
the end of the utterance end due to gradient vanishing. In order to reduce the sequence length processed
by GRUs, we sub-sample~\cite{bahdanau2014neural, chan2016listen} the hidden activations along the time domain
for every bi-direction GRU layer (see Fig.~\ref{fig:slu}). 
This allowed us to extract a representation roughly at a syllable level 
in a given utterance. On the other hand, this significantly reduced the
computational time for both training and prediction, which enables us to use  
bi-directional GRU for real time intent and/or domain classification. 
Given the encoder output, 
a max-pooling layer along the time axis is used to compress it into a fixed dimension vector. 
This is followed by a fully-connected
feed-forward layer. Finally, a softmax layer is used to compute the posterior
probability of intents or domains.

\section{Experiments}
\label{sec:experiments}

We train and evaluate our models on an in-house dataset containing VR spoken
commands collected for that purpose. The dataset is close in spirit 
to ATIS corpus~\cite{hemphill1990atis}. The dataset contains about 320 hours of
near field annotated data collected from a diverse set of more than 1000 de-identified speakers. It was recorded in two scenarios: scripted
and free speech. The scripted half is read speech with a fixed script. For the
free speech part, the participants were asked to achieve certain goal using any phrasing, then these utterances were
transcribed. Every utterance has transcription as well as meta information including a domain label and an intent label. After cleaning
data we extracted 5 domains and 35 distinct intents. Roughly 11,000 utterances,
totaling 10 hours audio, are used as the evaluation set. 

For all our experiments we used log-Mel filterbank features. We used an encoder-decoder architecture~\cite{sutskever2014sequence}. The encoder
is a 4 layer bidirectional GRU network with 256 units every layer. The output
was sub-sampled with a stride of 2 at every layer to reduce
the length of the representation. The decoder network
takes the output of
the encoder and aggregates them with a max-pooling layer and
puts it through a 1-layer feed-forward network (hidden size 1024) to produce
either the domain or intent class. The network was optimized with Adam~\cite{kingma2014adam} algorithm until convergence on the validation set.
We used the \emph{batch normalization}~\cite{ioffe2015batch} for every feed-forward connection to speed up the convergence of this network.

The baseline NLU model for our experiments was a recurrent network similar to
our model with an exception that we did not use sub-sampling.  
The input word was represented as one-hot vector and a trainable embedding was used.
This model is close to state of the art models and ones used in production~\cite{collobert2011natural, lai2015recurrent}. 
This network was evaluated in two regimes: using the transcript text,
and using the recognized text. The former shows the upper bound for our models and corresponds to the perfect speech recognition. The later 
regime was transcribed by a speech recognizer: the core part of AM is a
four-layer CD-HMM-LSTM network\cite{sak2014long} with 800 memory cells in each
layer, trained on the same 320 hours training data with a cross-entropy
criterion to predict posterior probabilities of 6,133 clustered
content-dependent 
states; the vocabulary size is 80,000 and the LM is trained on the transcribed
320 hours speech and interpolated with a large background LM which is trained on
other sources; there are roughly 200M n-grams in the LM. Our ASR system achieved 3.5\% word error
rate on the evaluation set. Each utterance from the evaluation set was recognized with this ASR and inputed to the baseline NLU network. In order to provide the
fair comparison we did not use any external data in both cases. No pretrained embeddings or dictionary look-up was used.

We also emulate the real-world situation where the input
to the SLU system is noisy. Both training and evaluation datasets were corrupted by convolving with
recorded room impulse responses (RIRs) whose T60 times ranges from 200ms to 1 second. Background noise
was added as well: for training data, the SNR ranges from 5 to 25dB, while for
evaluation data, the SNR ranges from 0 to 20dB. Every training utterance is
distorted 2 times by using different RIRs, sources of background noise and
SNRs. This results in a 600 hours noise-corrupted training set. Utterances in
evaluation set are only distorted once. Both the ASR system and our end-to-end 
system were retrained on this noise-corrupted training set. 
Due to the reverberation and relatively strong background noise, the ASR has considerably
higher word error rate (28.6\% WER). 

\section{Results and discussion}
\label{sec:results}
\begin{figure}[t]
    \centering
    \begin{subfigure}[b]{0.9\linewidth}
      \includegraphics[width=\linewidth]{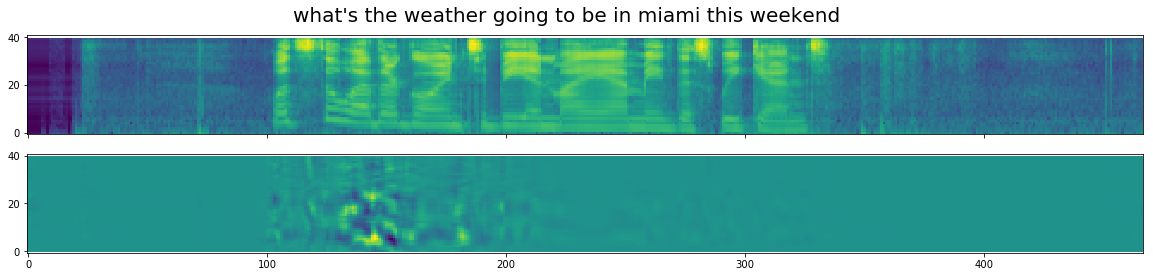}
    \end{subfigure}
    
    \begin{subfigure}[b]{0.9\linewidth}
      \includegraphics[width=\linewidth]{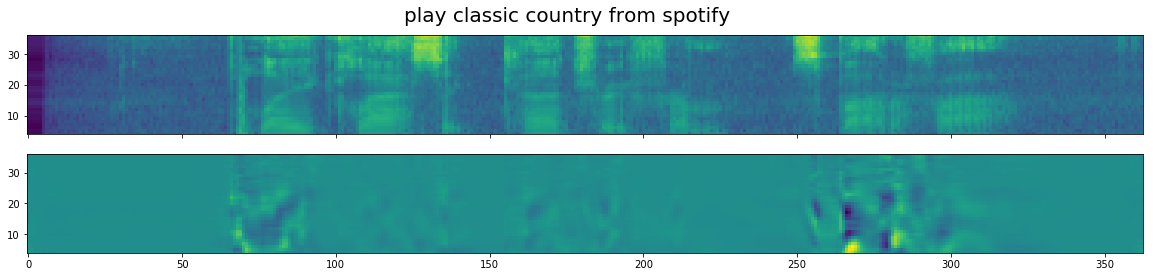}
    \end{subfigure}
    \caption{Filter-bank features and corresponding saliency maps for our
      speech-to-domain model. It can be seen that the model responds to 
    ``what's the weather'' part and ``play'' or ``spotify'' from the second example. Top image is ``weather'' domain, bottom image is 
    ``music'' domain.}
    \vspace{-2em}
    \label{fig:saliency}
\end{figure}

We first present results of domain classification. 
Five domains in this corpus are ``music,'' ``weather,'' ``news,'' ``sports,'' ``opt-in, opt-out.''
The results for domain recognition are summarized in Table~\ref{tab:domain}. The performance
in ``Transcript text'' row is evaluated on perfect transcriptions that
correspond to human transcription and such performance is not achievable with the current ASR
system. For qualitative analysis of our end-to-end model, 
we visualize the saliency map (gradients w.r.t. input) in Fig.~\ref{fig:saliency} for a selected
utterance. We notice that the position of the network response corresponds to meaningful 
positions in the utterance. As it can be seen in Table~\ref{tab:domain}, 
the accuracy is close to perfect on this dataset. Therefore we continue with a more challenging
task of the intent classification.
\begin{table}[htbp]
    \centering
    \caption{Results for domain classification. The first row corresponds to evaluation on clean transcripts, the maximum performance
      achievable with this model.}
    \label{tab:domain}
    \begin{tabular}{lr}
    \toprule
    {\bf Input} & {\bf Accuracy, \%} \\
    \midrule
    Transcript text & 99.2 \\
    Recognized text & 98.1 \\
    Audio           & 97.2 \\
    \bottomrule
    \end{tabular}
\end{table}

Examples of intents are 
``Learn about the weather in your location,'' ``Learn about the weather in different location,'' 
``Learn the results of a completed sports match, game, tournament,''
``Learn the status of an ongoing sports match, game, tournament.'' 
These classes are more fine grained and require deeper understanding  of a given utterance. 
For the ablation study we report the performance of
variations of our end-to-end model: initially, we use last-layer hidden activations at
the end of both left-to-right and right-to-left directions as the encoder output;
this is compared with using max pooling to aggregate all the hidden activations
in the last layers in Table~\ref{tab:intent}, which shows that using max-pooling
activations increases performance. 
We hypothesize that due to the long input speech
sequence (ranging from 100 to 1,000 input frames), the activations in the last time
step may not sufficiently summarize the utterance. This is contrary to the
NLU model using text as input, whereas using hidden activations from
the last time step usually suffice, as the input length is relatively small
(a majority of them are less than 20 words). We also observe that using batch normalization
(BN) in both encoder and decoder significantly improves the classification
accuracy, which indicates that our end-to-end model is difficult to optimize
due to the long input sequence. The results for the noise-corrupted data are
also reported in the bottom part of Table~\ref{tab:intent}. The performance of both models degrades
significantly. A potential advantage of our end-to-end approach to SLU is that
since the model is optimized under the same criterion, the model can be made
very compact; this is demonstrated in the same table by comparing the
number of parameters of neural networks in standard SLU system and our system.
Note that the parameters in the LM are not included, although it is usually two or
three magnitude more than AM in ASR. Due to the compactness of our end-to-end
model, we are able to predict the intent or domain with a real time factor
around 0.002, which makes the use of bi-GRU in our model possible for real time
applications.

\begin{table}[tbp]
    \centering
    \caption{Results for intent classification. As mentioned above, the first row is the maximum performance achievable with
      this model. We report the number of parameters for the models used to demonstrate that a much smaller end-to-end model
      achieves reasonable performance.}
    \label{tab:intent}
    \begin{tabular}{lrr}
    \toprule
    {\bf Input} & {\bf Parameters} & {\bf Accuracy, \%} \\
    \midrule
    \emph{Text input} \\
    Transcript         & 15.5M & 84.0 \\
    Recognized         & 15.5M & 80.9 \\
    \midrule
    \emph{Clean audio features} \\
    no BN, no max-pool & 0.4M & 71.3 \\
    no BN, max-pool    & 0.4M & 72.5 \\ 
    BN, max-pool       & 0.4M & 74.1 \\
    \midrule
    \emph{Noisy audio features} \\
    Recognized text    & 15.4M & 72.0 \\
    Audio              & 0.4M  & 64.9 \\
    \bottomrule
    \end{tabular}
\end{table}

With this work we hope to start a discussion on the topic of the audio-based SLU. 
Although, the end-to-end approach does not show superior performance, it provides a promising 
research direction. With significantly less parameters our system is able to reach 10\% relatively worse accuracy.
One of the future directions for this work is to encompass the slot filling task into this framework. It requires simultaneous prediction of a word
and a slot. This can be tackled with the attention-based networks. Other directions are to explore different architectures for the decoder, such
as using different pooling strategies, using deeper networks and incorporating convolutional transformations.


\ninept
\bibliographystyle{IEEEbib}
\bibliography{refs}

\end{document}